\documentclass[11pt]{article}
\usepackage{multicol}

\usepackage{amsmath,graphicx}
\usepackage{graphicx}
\usepackage{color}
\usepackage{placeins}
\usepackage{float}
\usepackage{tabularx,colortbl}
\graphicspath{{figures/Appendix/}}
\usepackage{amssymb} 
\usepackage{amsmath} 
\usepackage[breaklinks=true]{hyperref}
\usepackage{setspace}
\usepackage{cite}
\usepackage{subfigure}

\renewcommand{\vec}[1]{ \mbox{$\mathbf {#1}$}}

\newcommand{\vx}{\mbox{$\vec{x}$}}

\newcommand{\grad}{\nabla}

\newcommand{\MB}{\left[\begin{array}}
\newcommand{\ME}{\end{array}\right]}

\DeclareGraphicsExtensions{.pdf,.png,.jpg}

\begin{document}
\begin{center}
\textbf{Cosine Similarity Measure According to a Convex Cost Function}

{Osman Gunay, Cem Emre Akbas, A. Enis Cetin}

{Department of Electrical and Electronic Engineering
	\\  Bilkent University, 06800 Bilkent,  \\
	Ankara, Turkey\\
E-mail:akbas@ee.bilkent.edu.tr, osman@ee.bilkent.edu.tr, cetin@bilkent.edu.tr\\}

\end{center}
%
%
\begin{abstract}
In this paper, we describe a new vector similarity measure associated with a convex cost function. Given two vectors, we determine the surface normals of the convex function at the vectors. The angle between the two surface normals is the similarity measure. Convex cost function can be the negative entropy function, total variation (TV) function and filtered variation function. The convex cost function need not be differentiable everywhere. In general,  we need to compute the gradient of the cost function to compute the surface normals. If the gradient does not exist at a given vector, it is possible to use the subgradients and the normal producing the smallest angle between the two vectors is used to compute the similarity measure. 
\end{abstract}
%
%
\section{Introduction}
Inner product of two vectors is the basis of many big data analysis, machine learning and signal processing algorithms~\cite{ref1}. For example,
the cosine similarity between two vectors $\textbf{x}_1$ and $\textbf{x}_2$  is computed using the inner product of the two vectors divided by the $\ell_2$-norms of the vectors: 
\begin{equation}
\label{eq:cosine}
\cos (\textbf{x}_1, \textbf{x}_2) = \frac{\langle\textbf{x}_1 , \textbf{x}_2\rangle}{\|\textbf{x}_{1}\|  \|\textbf{x}_{2}\|},
\end{equation}

In this article, we want to determine the similarity of two vectors according an associated convex cost function $f$. In Figure 1, the main idea behind the new cost measure is graphically described. Tangent lines and the surface normals at $\textbf{x}_1$, $f(\textbf{x}_1)$ and $\textbf{x}_2$ and $f(\textbf{x}_2)$ are determined. We propose a similarity measure that can be defined as the cosine similarity between the surface normals of the two vectors $\textbf{x}_1$ and $\textbf{x}_2$ on the convex cost function $f$ as follows:
\begin{equation}
C(\vx_1, \vx_2) = \langle \vec{e}_1, \vec{e}_2 \rangle
\end{equation}
where $\vec{e}_1$ and $\vec{e}_2$ are the unit surface normal vectors of the convex cost function $f$ at  $\textbf{x}_1$, and $\textbf{x}_2$, respectively. We call the cosine measure $C(\vx_1, \vx_2)$ Bregman angle between $\textbf{x}_1$, and $\textbf{x}_2$.

This new measure is inspired by the well-known Bregman divergence~\cite{ref2,ref3,ref4,ref5}. which is based on the surface tangent of the cost function. The Bregman divergence $D(\vx_1,\vx_2)$ between the two vectors $\textbf{x}_1$ and $\textbf{x}_2$ is the ``vertical'' distance between the cost function $f$ and the tangent line at $\textbf{x}_2$ evaluated at the vector $\textbf{x}_1$: 
\begin{equation}
D(\vx_1,\vx_2) = f(\vx_1)-f(\vx_2)-\grad f(\vx_2)^T(\vx_1-\vx_2)
\end{equation}
For example, when $f(\vx) = \|\vx\|^2$ then the Bregman divergence reduces to Euclidian or the square distance between the two vectors, i.e., $D(\vx_1,\vx_2) = \|\vx_1-\vx_2\|^2$. 

\begin{figure}[H]
\centering
  \centering
  \includegraphics[width=.9\linewidth]{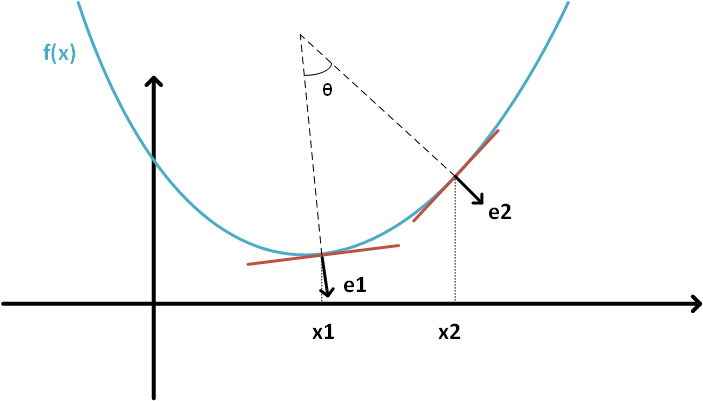}
  \caption{The angle between e1 and e2 is the similarity value between the two vectors x1 and x2.}
  \label{fig:alg}
\end{figure}
\section{Bregman Angle Similarity Measure}
For a convex function $f(x)$ the unit surface normal is defined as:
\begin{equation}
\vec{e} =\frac{[\nabla f(\vx),~~-1]}{||[\nabla f(\vx),~~-1]||}
\end{equation}
In the next subsection we use the surface normals of the convex function to construct vector similarity measures.
\subsection{Similarity Measure Based on Surface Normals}
The general form of the proposed similarity measure based on surface normals can be defined as follows:
\begin{equation}
C(\vx_1, \vx_2) = \frac{\langle\nabla f(\vx_1),\nabla f(\vx_2) \rangle+1}{\sqrt{\langle\nabla f(\vx_1),\nabla f(\vx_1) \rangle+1}\sqrt{\langle\nabla f(\vx_2),\nabla f(\vx_2) \rangle+1}}
\end{equation}

When the cost function is the well-known negative entropy $f(\vx) =  \sum_{i} \vx(i)\log(\vx(i))$ the surface normals are given by:
\begin{equation}
\vec{E}_1 = \left[\frac{\partial f(\vx_1)}{\partial\vx_1(1)},\cdots,\frac{\partial f(\vx_1)}{\partial\vx_1(N)}, -1 \right]=\left[\log(\vx_1(1))+1, \cdots, -1 \right]
\end{equation}
\begin{equation}
\vec{E}_2 = \left[\frac{\partial f(\vx_2)}{\partial\vx_2(1)},\cdots,\frac{\partial f(\vx_2)}{\partial\vx_2(N)}, -1 \right]=\left[\log(\vx_2(1))+1, \cdots, -1 \right]
\end{equation}
and unit normals are:
\begin{equation}
\vec{e}_1 =\frac{\vec{E}_1}{||\vec{E}_1||}
\end{equation}
\begin{equation}
\vec{e}_2 =\frac{\vec{E}_2}{||\vec{E}_2||}
\end{equation}

The cosine similarity between the vectors is then defined as follows:
\begin{equation}
C(\vx_1, \vx_2) = \frac{\sum_i{(\log(\vx_1(i))+1)(\log(\vx_2(i))+1)}+1}{\sqrt{\sum_i{(\log(\vx_1(i))+1)^2}+1}\sqrt{\sum_i{(\log(\vx_2(i))+1)^2}+1}}
\label{Equation:8}
\end{equation}

Since the entropy function is only defined for positive values we can use the modified entropy functional introduced in~\cite{kose} to account for non-negative values:
\begin{equation}
f(\vx)=\sum_i{\left(|\vx(i)|+\frac{1}{e}\right)\log\left(|\vx(i)|+\frac{1}{e}\right) + \frac{1}{e}}
\end{equation}

For this case the Bregman angle measure can be obtained from the following surface normals:
\begin{equation}
\vec{E}_1 = \left[\mbox{sign}(\vx_1(1))\left(\log\left(|\vx_1(1)|+\frac{1}{e}\right)+1\right), \cdots, -1 \right]
\end{equation}
\begin{equation}
\vec{E}_2 = \left[\mbox{sign}(\vx_2(1))\left(\log\left(|\vx_2(1)|+\frac{1}{e}\right)+1\right), \cdots, -1 \right]
\end{equation}

A well-known convex cost function is the total-variation (TV) function:
\begin{equation}
TV(\vx) = \sum_{i}^{N} |x_{i+1} - x_i|
\end{equation}
For the TV function the surface normal vector $SN(TV)$ is given by
\begin{equation}
SN(TV) = \left[\frac{\partial TV}{\partial x_1}, \frac{\partial TV}{\partial x_2}, \dots, \frac{\partial TV}{\partial x_N}, -1\right]
\end{equation}
which is equal to
\begin{equation}
[(\mbox{sign}(x_2 - x_1)), (\mbox{sign}(x_2 - x_1) - \mbox{sign}(x_3 - x_2)), \dots, (\mbox{sign}(x_{N} - x_{N-1}), -1]
\end{equation}
where $\mbox{sign}(.)$ is the signum function.
We can easily construct a vector similarity measure from the above vector. It turns out that we get the best experimental results using the TV function.

Similarly for $f(\vx) = \|\vx\|^2$ the distance function becomes:
\begin{equation}
C(\vx_1, \vx_2) = \frac{\sum_i{4\vx_1(i)\vx_2(i)}+1}{\sqrt{\sum_i{4\vx_1(i)^2}+1}\sqrt{\sum_i{4\vx_2(i)^2}+1}}
\end{equation}
When we remove the last entry from the surface normals the Bregman cosine similarity becomes the ordinary cosine similarity.

\begin{figure}[!htb]
\centering
\subfigure[Distribution of samples]{
  \centering
  \includegraphics[width=.45\linewidth]{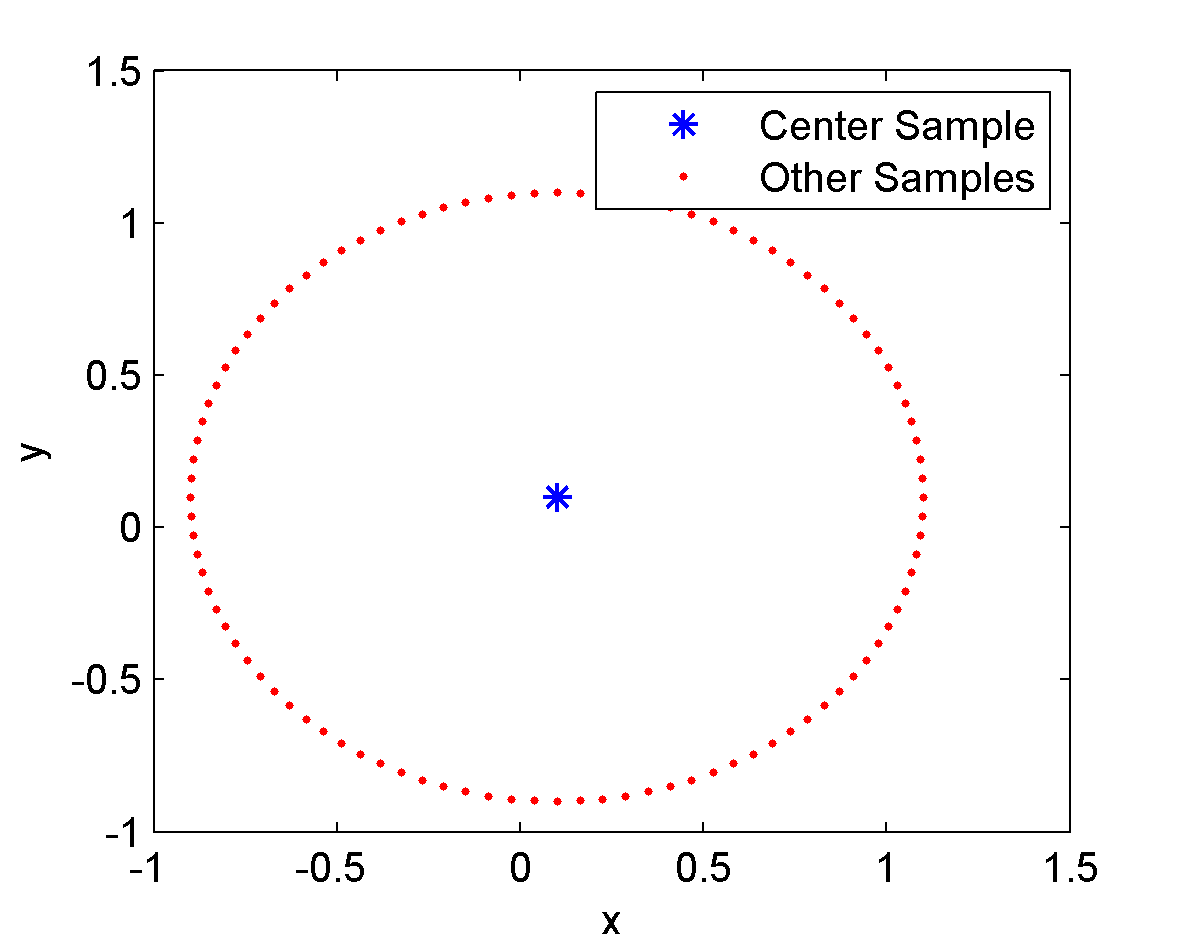}
}
\subfigure[Distance/Similarity measures]{
  \centering
  \includegraphics[width=.45\linewidth]{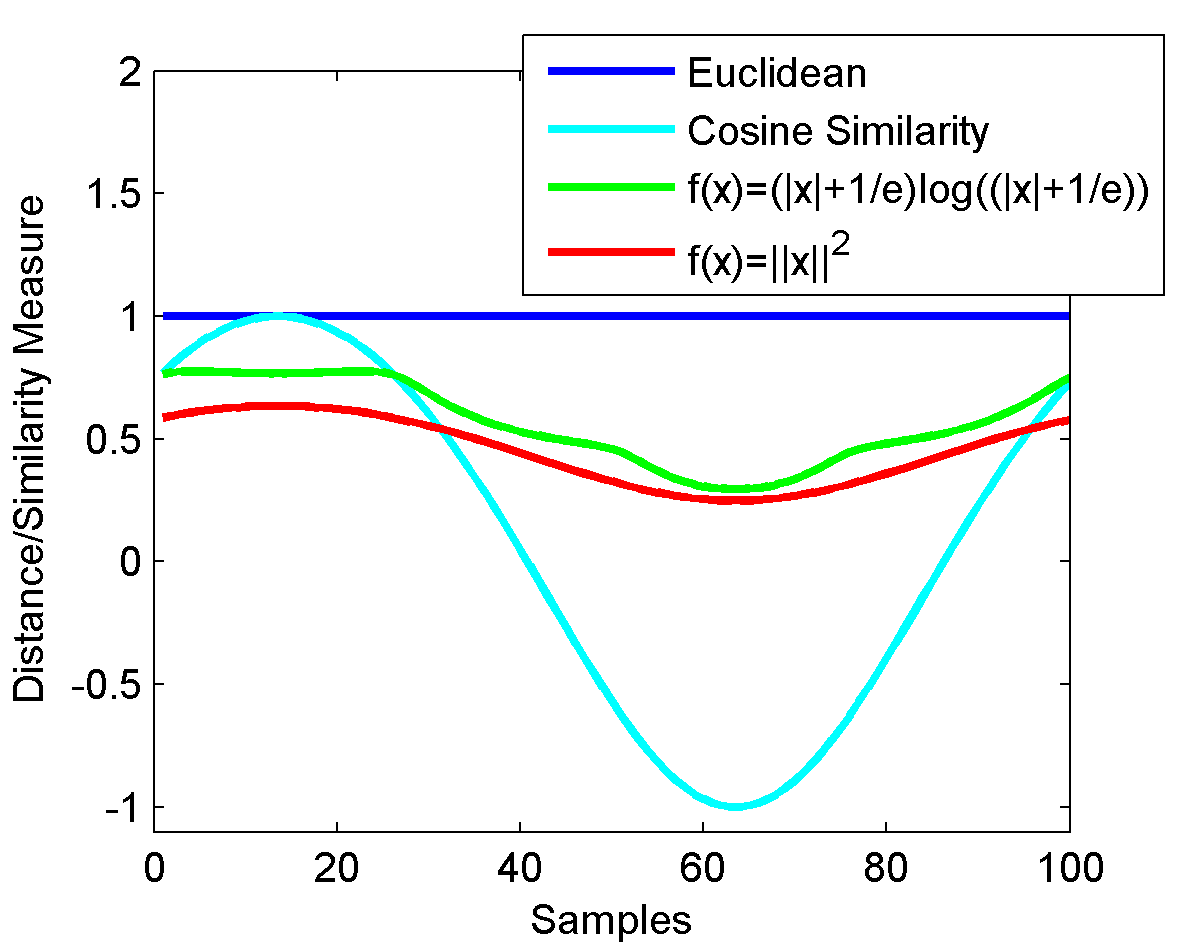}
}
\caption{Distance similarity measure for concentric distribution of samples.}
\label{fig:test1}
\end{figure}

\begin{figure}[!htb]
\centering
\subfigure[Distribution of samples]{
  \centering
  \includegraphics[width=.45\linewidth]{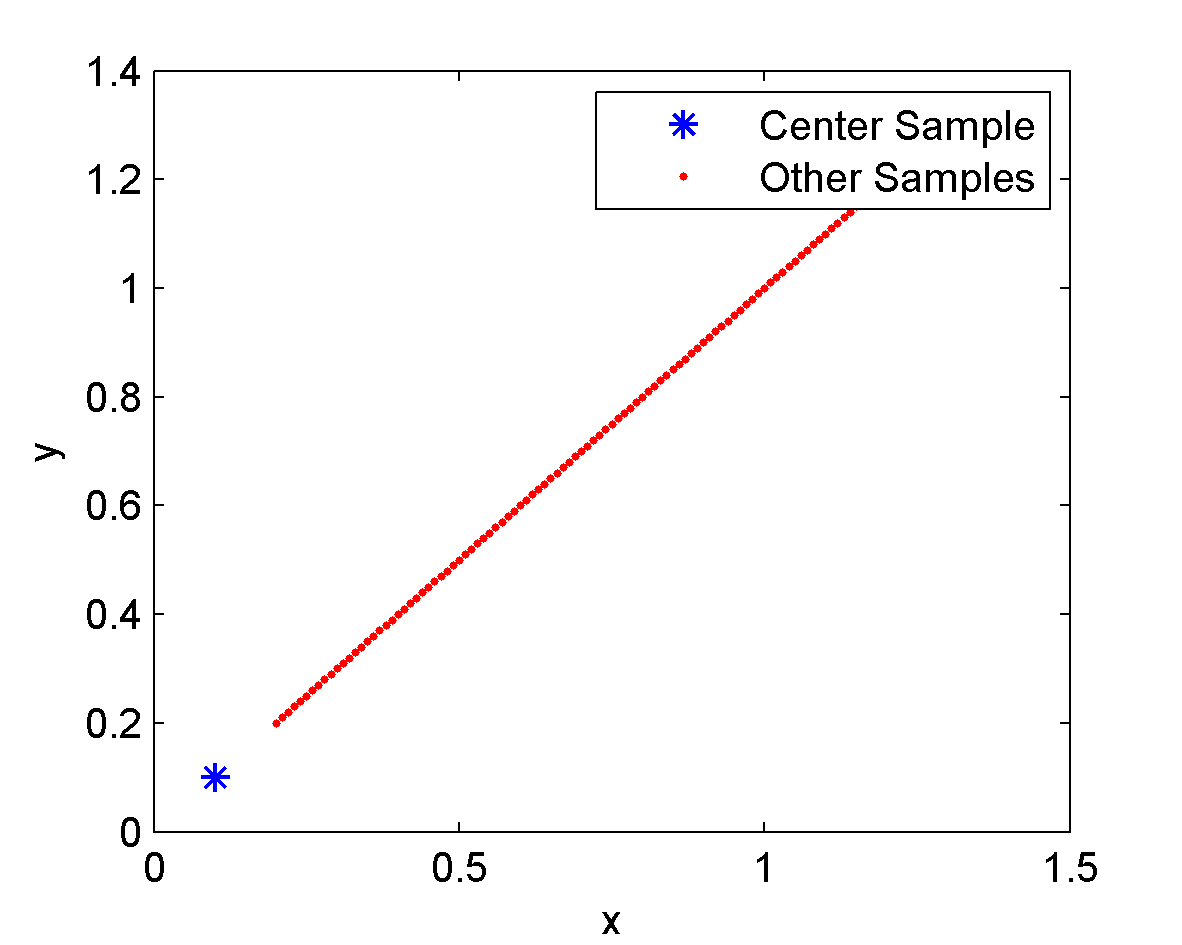}
}
\subfigure[Distance/Similarity measures]{
  \centering
  \includegraphics[width=.45\linewidth]{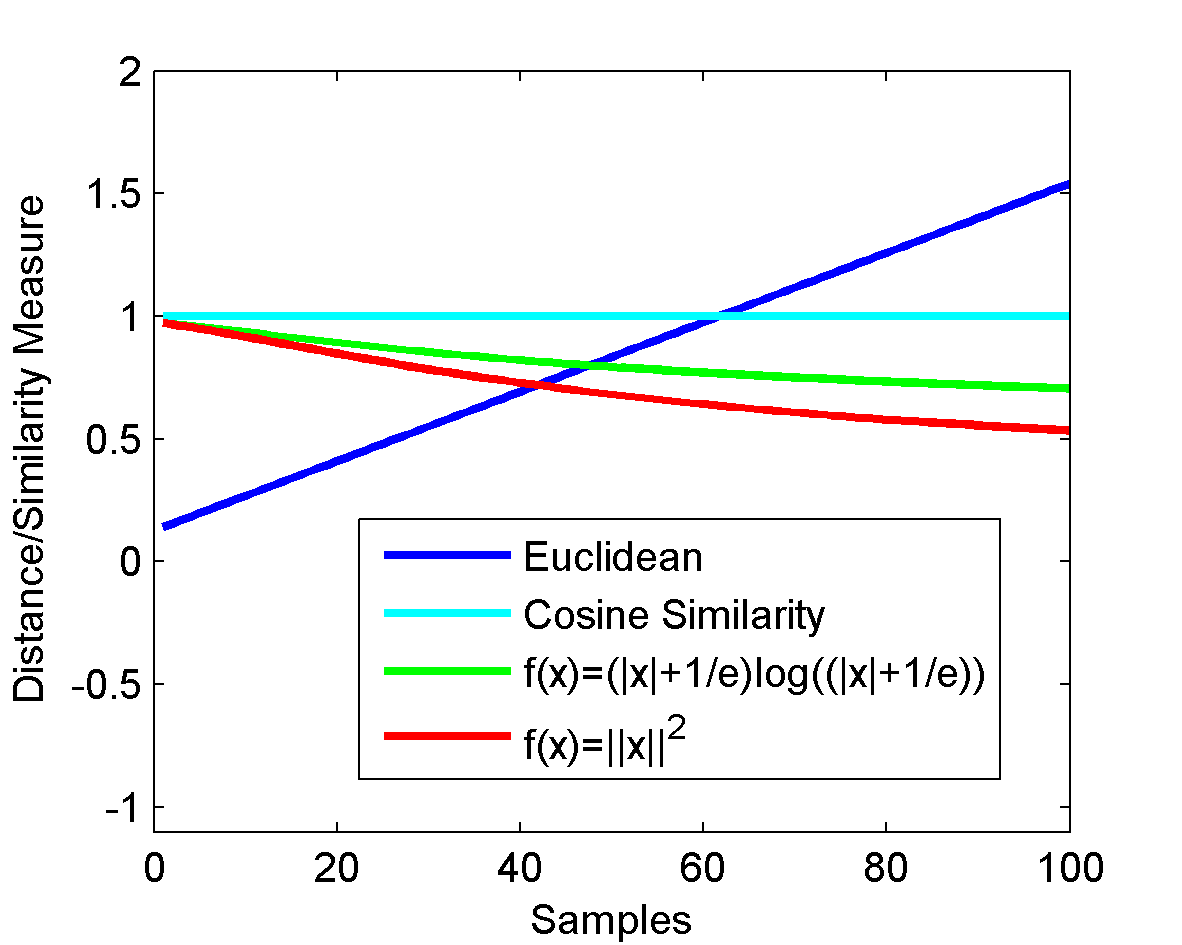}
}
\caption{Distance similarity measures for linear distribution of samples.}
\label{fig:test2}
\end{figure}

In Figure~\ref{fig:test1} and~\ref{fig:test2} some examples are shown to compare the proposed similarity measure for two extreme cases of sample distributions. When the samples are defined over a circle the Euclidean distance is same for all samples but cosine similarity and Bregman angle can distinguish between samples at different angles according to the center sample. When the samples are defined on a line usual cosine similarity cannot separate the samples but proposed Bregman angle measure still works.

\subsection{Similarity Measure Based on Surface Tangents}
When surface tangents are used instead of surface normals the similarity measure reduces to:
\begin{equation}
C_t(\vx_1, \vx_2) = \frac{\langle\nabla f(\vx_1),\nabla f(\vx_2) \rangle}{\sqrt{\langle\nabla f(\vx_1),\nabla f(\vx_1) \rangle}\sqrt{\langle\nabla f(\vx_2),\nabla f(\vx_2) \rangle}}
\end{equation}

Bregman distance uses surface tangents of the convex cost function.  Therefore we can also use the surface tangents to define another cosine similarity measure. Given two vectors $\textbf{x}_1$ and $\textbf{x}_2$, we compute the gradient vectors $t_1$ and $t_2$ of the cost function $f$ at  $\textbf{x}_1$ and $\textbf{x}_2$ and the angle between $t_1$ and $t_2$ is the cosine similarity measure.

For the negative entropy function the vector similarity measure becomes:
\begin{equation}
C_t(\vx_1, \vx_2) = \frac{\sum_i{(\log(\vx_1(i))+1)(\log(\vx_2(i))+1)}}{\sqrt{\sum_i{(\log(\vx_1(i)))^2}+1}\sqrt{\sum_i{(\log(\vx_2(i))+1)^2}}}
\label{Equation:9}
\end{equation}
This is similar to the Eq. (8) but the dimension of the inner product is smaller than Eq. (8).

When the cost function is the Euclidean distance function $C_t$ becomes the same as the ordinary cosine similarity vector.

\section{Experimental Results}
In the first experiment we compare the performance of the Bregman angle measure with cosine similarity measure on a gesture phase segmentation dataset. The gesture phase segmentation data set~\cite{Dataset} was made available by UC Irvine Machine Learning Repository. The data set contains 5 classes and 1747 gesture phase data each having 18 attributes. In this paper, simulations are carried out using the first two classes which contain total of 202 instances.

First, input vectors are multiplied by $10^7$ in order to improve classification performance of similarity measures. In all simulation studies, we have a leave-one-out strategy. The size of the test set is one and the training set contains the remaining data. The test set is circulated to cover all instances. 2 class 1-nearest neighbor classification is performed using the new similarity measure (\eqref{Equation:8}) and cosine similarity measures. Classification accuracies is given in Table 1.

In this data set the tangential similarity function described in (\eqref{Equation:9}) produces slightly lower results than the surface normal based similarity measure.

\begin{table}[ht!]
	\centering
	\caption{Classification accuracies (Percentage) for the 2 class 1-nearest neighbor classification with 2 different similarity measures. The last row is the ordinary cosine similarity measure.}
	\vspace{0.2cm}
	\label{table:1}
	\begin{tabular}{|c|c|c|c|c|c|c|c|c|c|}
	\hline			
	\textbf{Similarity/Distance Measure} & 	\textbf{Classification Accuracy} \\\hline
	Bregman Angle  (negative entropy)   &	\% 97.5 \\\hline
		Bregman Angle (TV)    &	\% 99.0 \\\hline
	Cosine Similarity &	\% 98.0 \\\hline
    \end{tabular}
\end{table}
	
As shown in Table \ref{table:1}, classification accuracy of our new similarity measure (\eqref{Equation:8}) is almost the same as classification accuracy of the cosine similarity measure. The TV function based similarity measure produces the best results in this dataset.

In the second experiment we used the KTH-TIPS database that contains 800 images for 10 different classes of colored textures~\cite{kth1}. We use half of the images for each class as the training set and the rest as the test set. We use 1-neighbour knn classifier and four diffierent distance/similarity measures. To extract features from the images we used the dual-tree complex wavelet transform (DT-CWT) as texture features and histograms in HSV color space as color features. Dual-tree complex wavelet transform tree, is recently developed to overcome the shortcomings of conventional wavelet transform, such as shift variance and poor directional selectivity~\cite{cw1}. To obtain wavelet features we divide images into four non-overlapping blocks and calculate the energies and variances of six different subbands (oriented at +/-15, +/-45, +/- 75) for each block. The combined feature vectors of all blocks are used as the texture feature of the image. The results for this test are in shown Table~\ref{table:2}. From the results we see that proposed measure have similar performance to Euclidean and cosine similarity measures.

\begin{table}[ht!]
	\centering
	\caption{Classification accuracies (Percentage) for KTH-TIPS dataset.}
	\vspace{0.2cm}
	\label{table:2}
	\begin{tabular}{|c|c|}
	\hline			
	\textbf{Similarity/Distance Measure} & 	\textbf{Classification Accuracy} \\\hline
	Euclidean Distance    		 &	(381/400)	\% 95.25 \\\hline
	Cosine Similarity     		 &	(380/400)	\% 95.0 \\\hline
	Bregman Angle (Entropy)      &	(379/400)	\% 94.75 \\\hline
	Bregman Angle (l2-norm)      &	(380/400)	\% 95.0 \\\hline
    \end{tabular}
\end{table}

\section{Conclusion} 
In this paper, we introduced new vector similarity measures based on a convex cost function.  The angle between the two surface normals or surface tangents are used to construct the similarity measures. When the cost function is the ordinary Euclidean function the surface tangent based similarity measure reduces to the ordinary cosine measure. It is experimentally observed that TV function based vector similarity measure produces the best results in a dataset containing human gesture data.

\bibliographystyle{IEEETran}
\bibliography{refer}

\end{document}